\documentclass[conference]{IEEEtran}
\IEEEoverridecommandlockouts

\usepackage{cite}
\usepackage{amsmath,amssymb,amsfonts}
\usepackage{algorithmic}
\usepackage{graphicx}
\usepackage{textcomp}
\usepackage{xcolor}
\usepackage{tabularx}
\usepackage{booktabs} 
\usepackage{caption}
\usepackage{array}
\usepackage{enumitem}
\def\BibTeX{{\rm B\kern-.05em{\sc i\kern-.025em b}\kern-.08em
    T\kern-.1667em\lower.7ex\hbox{E}\kern-.125emX}}

\begin{document}

\title{SQ-DM: Accelerating Diffusion Models with \\Aggressive Quantization and Temporal Sparsity \\
}

\author{\IEEEauthorblockN{Zichen Fan\textsuperscript{*}\textsuperscript{\ddag}, Steve Dai\textsuperscript{\dag}\textsuperscript{\#}, Rangharajan Venkatesan\textsuperscript{\dag}, Dennis Sylvester\textsuperscript{*}, Brucek Khailany\textsuperscript{\dag}}
\IEEEauthorblockA{
\textsuperscript{*}\textit{University of Michigan, Ann Arbor, MI}, \hspace{10pt}\textsuperscript{\dag}\textit{NVIDIA, Santa Clara, CA}\\
\textsuperscript{\ddag}zcfan@umich.edu,\hspace{10pt} \textsuperscript{\#}sdai@nvidia.com}
}

\maketitle

\begin{abstract}

Diffusion models have gained significant popularity in image generation tasks. 
However, generating high-quality content remains notably slow because it requires running model inference over many time steps. 
To accelerate these models, we propose to aggressively quantize both weights and activations, while simultaneously promoting significant activation sparsity. 
We further observe that the stated sparsity pattern varies among different channels and evolves across time steps. 
To support this quantization and sparsity scheme, we present a novel diffusion model accelerator featuring a heterogeneous mixed-precision dense-sparse architecture, channel-last address mapping, and a time-step-aware sparsity detector for efficient handling of the sparsity pattern. 
Our 4-bit quantization technique demonstrates superior generation quality compared to existing 4-bit methods. 
Our custom accelerator achieves \(6.91\times\) speed-up and \(51.5\%\) energy reduction compared to traditional dense accelerators.


\end{abstract}


\section{Introduction}

Diffusion models generate realistic contents by sequentially denoising data from random noise.
They have emerged as the cornerstone of generative artificial intelligence (GenAI) for tasks such as image generation~\cite{balaji2022ediff}, video generation~\cite{blattmann2023align}, and even weather prediction~\cite{mardani2023generative}.
However, generating high-quality contents is notably slow because it requires repeatedly evaluating a large and complex neural network model over many time steps.
Its complexity increases with higher resolution images and more advanced models, leading to large computation and memory overheads.
Therefore, it is crucial to accelerate diffusion models to enable efficient deployment of real-world applications.

Quantization and sparsity serve as the two pivotal techniques driving dramatic improvements in deep neural network (DNN) performance over the last decade.
Quantization enables reducing the precision of the weights and activations of a DNN, typically from 32-bit floating-point to low bitwidth formats like 4-bit integers. 
Sparsity refers to the practice of pruning less important weights and activations, such as zeroing out the two smallest values for every four adjacent values in a tensor. 
In isolation or combination, quantization and sparsity can lead to substantial improvements in compute and memory efficiency of DNN execution with smaller model sizes, faster computations, and lower power consumption.


While quantization and sparsity remain the first-order considerations when accelerating diffusion models, these techniques fail to work out-of-the-box due to the unique characteristics of these models.
First, quantization error accumulates over time steps. 
Even small error in a single evaluation (time step) of the model becomes compounded over the many model evaluations required to generate a single sample (one image).
Second, activation distributions in diffusion models vary across layers and time steps.
This makes it difficult to apply a uniform quantization scheme over the entire generation process.
Third, the data values within diffusion models tend to be extremely dense.
The observed level of density precludes the effective use of sparse DNN accelerators for executing these models more efficiently.

\begin{figure}[t]
    \centering
    \includegraphics[trim={0 0 0 10pt}, clip, width=0.99\linewidth]{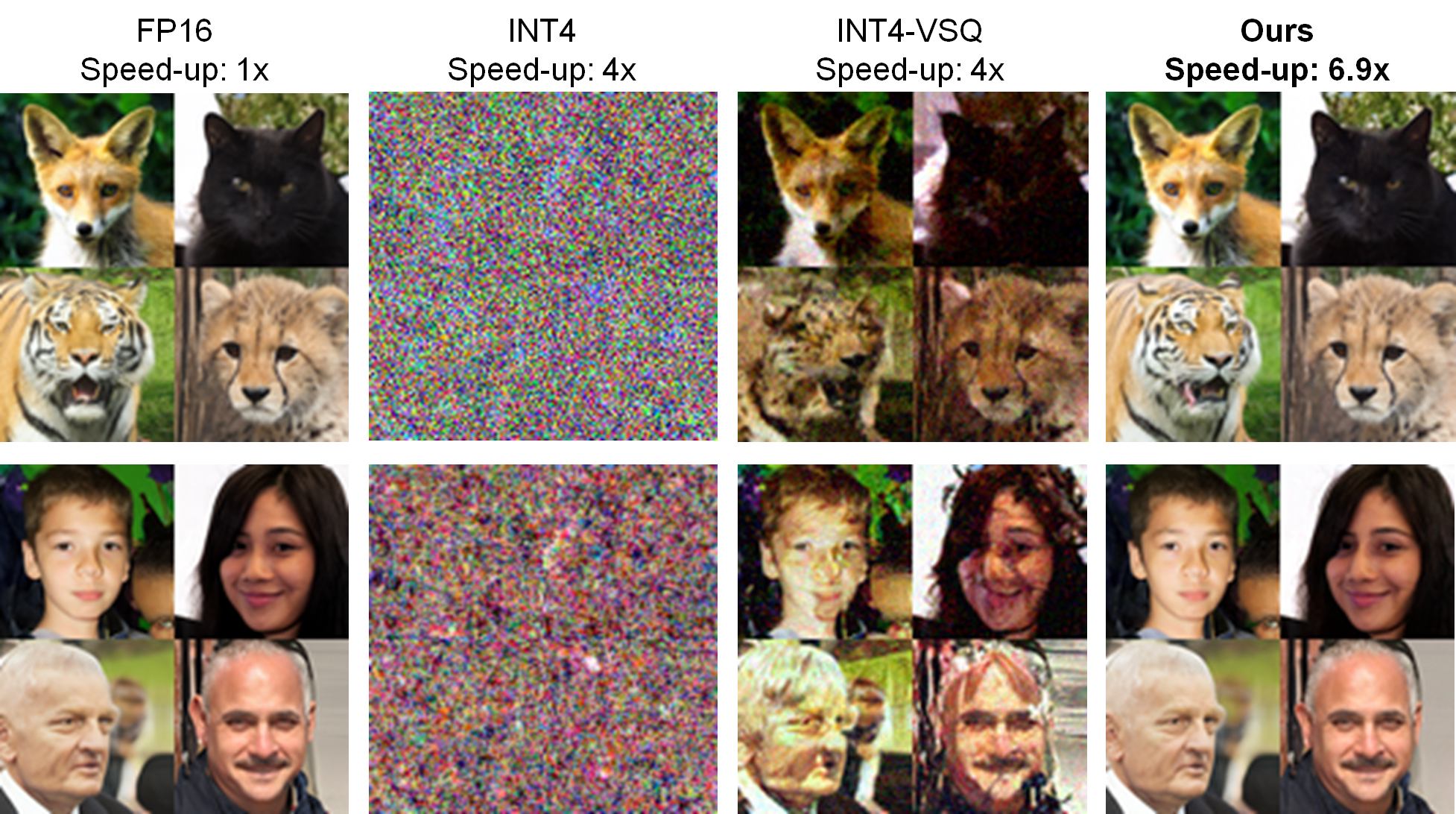}
    \caption{Generated images and achieved speed-up for different data formats and quantization techniques. }
    \label{fig:image_example}
\end{figure}

In fact, we observe severe quality degradation when applying existing 4-bit data formats to the diffusion model.
As shown in Figure~\ref{fig:image_example}, existing 4-bit formats such as \texttt{INT4} and \texttt{INT4-VSQ} result in obvious and unacceptable image quality degradation.
In this paper, we propose novel software-hardware co-design techniques to accelerate diffusion models to achieve state-of-the-art generation quality at significantly reduced hardware costs.
Our contributions are as follows:
\begin{itemize}[leftmargin=13pt]
    \item We are the first to aggressively quantize both weights and activations of diffusion models to 4-bit while promoting significant activation sparsity in these models.
    \item We propose a method to fully exploit the temporal nature of the sparsity during diffusion model sampling to maximize the efficiency of image generation.
    \item We present a new mixed-precision dense-sparse accelerator featuring channel-last addressing and temporal sparsity detection to effectively handle the sparsity pattern.
\end{itemize}
The rest of the paper is structured as follows:
Section~\ref{sec:prelim} provides preliminaries on diffusion models and discusses relevant work in quantizing and sparsifying these models.
Section~\ref{sec:software} presents optimizations we propose to improve diffusion model efficiency.
Section~\ref{sec:hardware} details our accelerator implementation to support those optimizations.
Section~\ref{sec:conclusions} summarizes our findings and possible future directions.


\begin{figure*}[htbp]
    \centering
    \includegraphics[width=0.85\linewidth]{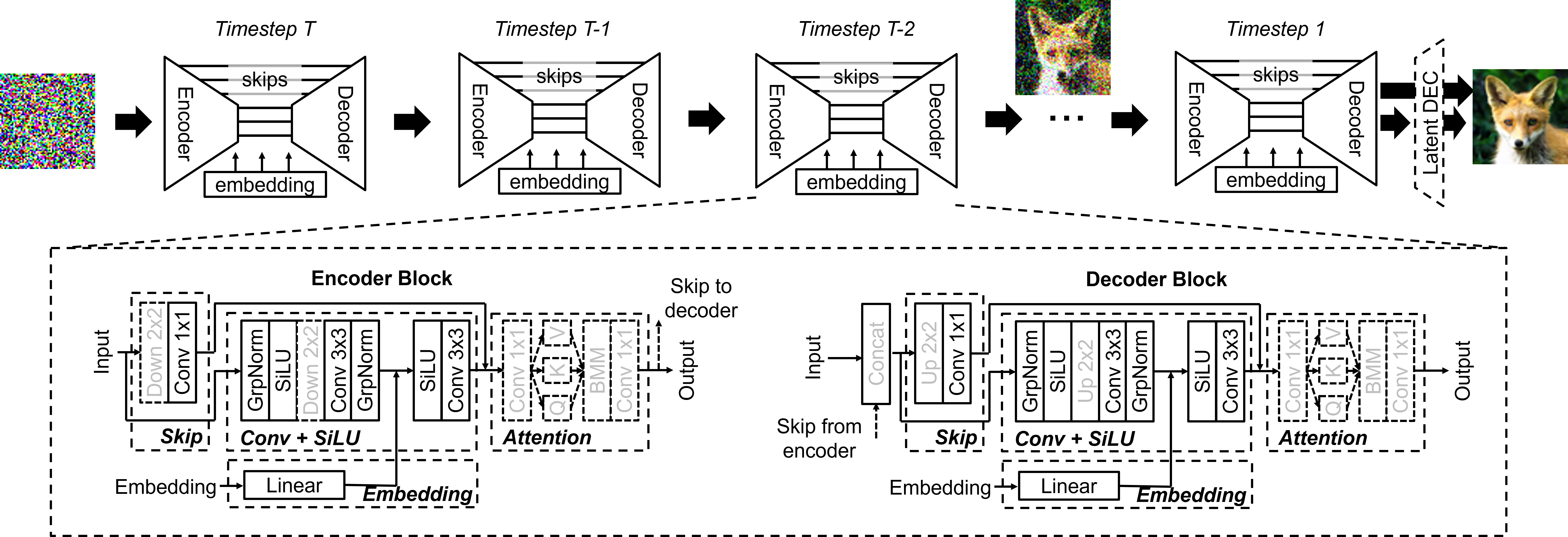}
    \caption{Execution process and model architecture of EDM~\cite{karras2022elucidating, karras2023analyzing}. }
    \label{fig:edm_model}
\end{figure*}

\section{Preliminaries}
\label{sec:prelim}

For the purpose of this work, we use Elucidated Diffusion Models, EDM \cite{karras2022elucidating} and EDM2 \cite{karras2023analyzing}, as our baseline diffusion models. 
EDM is the strongest baseline available today in terms of training and sampling techniques because it is built by extensively exploring the design space for training and sampling for this type of model.
More importantly, it is considered the state-of-the-art for its generation speed and quality, and thus very challenging to accelerate while maintaining good quality.
EDM serves as the backbone model for applications such as text-to-image generation~\cite{balaji2022ediff} and weather prediction~\cite{mardani2023generative}, and generalizes to the family of convolution-based diffusion models.

Figure~\ref{fig:edm_model} illustrates the execution process and model architecture of EDM. 
Like most convolution-based diffusion models, EDM employs a U-Net based architecture with an encoder to capture features of the noisy input and a decoder to reconstruct the denoised output.
As shown in Figure~\ref{fig:edm_model}, EDM's architecture mainly consists of four types of layers: \textit{Skip}, \textit{Conv+SiLU}, \textit{Embedding}, and \textit{Attention}.
The \textit{Skip} block manages the skip connections from the encoder to the decoder, enabling information propagation from encoder to decoder and earlier layers to deeper ones. 
The \textit{Embedding} linear layer transmits embedding information, including noise scheduling and labels (for conditional generation scenarios). 
The \textit{Attention} block, present in specific layers (e.g., \texttt{enc.16x16\_block\_1} in the EDM1 model for CIFAR-10), implements an image self-attention mechanism to capture global context. 
Typically, dozens to hundreds of time steps (a.k.a. encoder-decoder model evaluations) must be executed to generate an image from random noise, as each time step returns a slightly denoised image from the previous time step.


\subsection{Quantization}
\label{sec:quant}
DNN quantization converts model weights and activations from high-precision floating-point down to low-precision data formats. 
Specifically, uniform symmetric quantization of a tensor \(\mathbf{X}\) is expressed as $\mathbf{\hat{X}} =\mathrm{round}({\mathbf{X}}/s_x)$ where $s_x = \max(|\mathbf{X}|)/q_{\text{max}}$.
Here \(\mathbf{X}\) is the original tensor, \(\mathbf{\hat{X}}\) is the quantized tensor, and \(s_x\) is the quantization scaling factor for \(\mathbf{X}\). 
$q_{\text{max}}$ represents the largest quantized value.
The $\text{max}$ operator can be taken at different granularity of \(\mathbf{X}\), such as over the entire tensor, across each channel, or for each vector~\cite{dai2021vs}.
With scaled quantization, data will be stored and computed in the quantized format.
Computation results need to be rescaled using the scale factors \{\(s_x\)\} back to the model's original dynamic range.


Different techniques have been proposed to quantize diffusion model. 
PTQ4DM~\cite{shang2023post} and Q-diffusion~\cite{Li_2023} successfully quantize diffusion models to 8-bit. 
Several follow-up works \cite{he2023ptqd, yang2023efficient, sun2024tmpq,huang2024tfmq} further improve their generation quality. 
Recently, SVDquant  demonstrates 4-bit quantization on diffusion models using activation smoothing and low-rank decomposition~\cite{li2024svdqunat}. 
However, it requires high-precision (FP16) low-rank branches and special kernel fusion tricks to alleviate the overhead of these branches. 
In contrast, we directly quantize both the original weights and activations to 4-bit with little quality degradation without smoothing or using any high-precision floating-point components.
Our quantization technique also promotes sparsity simultaneously to enable further acceleration.

\subsection{Sparsification}
Sparsity in DNNs refers to the presence of zero-valued weights or activations, which can be exploited to reduce computational complexity and memory usage. 
In addition to quantization, we can prune less important weights and activations to create a sparse model that requires fewer resources.
In fact, current-generation Tensor Cores provide support for structured weight sparsity that exploits a 2:4 (50\%) sparsity pattern that enables twice the math throughput of dense matrix multiplications~\cite{mishra2021accelerating}.

Previous works such as \cite{fang2023structural} and \cite{wang2024sparsedm} have demonstrated that implementing structured weight sparsity in diffusion models can achieve the promised nearly 50\% reduction in computation.
These works leverage sparsity-aware finetuning and report modest degradation in model quality. 
Orthogonal to these efforts, we focus on activations instead of weights and promote sparsity that is not inherently present in the original model. 
We are able induce an average activation sparsity of 65\% that leads to approximately 52\% reduction in computational cost.
Activation sparsity can be combined with weight sparsity to enable additional efficiency.

\section{Diffusion Model Optimizations}
\label{sec:software}

To understand the landscape of quantizing diffusion models, we evaluate existing quantization techniques for EDM across various datasets (CIFAR-10 \cite{krizhevsky2009learning}, AFHQv2\cite{choi2020stargan}, FFHQ\cite{karras2019style} and ImageNet\cite{deng2009imagenet}). 
Following EDM and previous work, we use Fréchet Inception Distance (FID)~\cite{heusel2017gans} to measure the quality of generated images: lower FID means better image quality. 
In our experiments, we generate 50,000 images in CIFAR-10, AFHQv2, FFHQ and 10,000 images in ImageNet to calculate the FID scores. 
Table~\ref{table:normal_quant} shows the results of the models in full-precision (FP32), half-precision (FP16), as well as variants of the most competitive and commonly accepted 8-bit and 4-bit formats.
Here we use the same format for both weights and activations uniformly across the entire model.

\begin{table}[t]
    \centering
    \caption{FID comparison of different models across datasets with various existing quantization formats.}
    \label{table:normal_quant}
    \begin{tabularx}{0.5\textwidth}{>{\centering\arraybackslash}m{1.9cm}|>{\centering\arraybackslash}X|>{\centering\arraybackslash}X|>{\centering\arraybackslash}X|>{\centering\arraybackslash}X}
        \toprule
        \textbf{Model, Dataset} & \textbf{EDM1, CIFAR-10} & \textbf{EDM1, AFHQv2} & \textbf{EDM1, FFHQ} & \textbf{EDM2, ImageNet} \\
        \midrule
        FP32 & 1.85 & 2.08 & 2.46 & 4.47 \\
        \hline
        FP16 & 1.85 & 2.08 & 2.48 & 4.47 \\
        \hline
        INT8 & 12.60 & 14.46 & 20.34 & 8.78 \\
        \hline
        MXINT8 \cite{rouhani2023microscaling} & 1.93 & 2.08 & 2.63 & 4.48 \\
        \hline
        INT4 & 136.5 & 289.3 & 244.8 & 346.2 \\
        \hline
        INT4-VSQ \cite{dai2021vs} & 15.30 & 18.35 & 30.52 & 20.27 \\
        \bottomrule
    \end{tabularx}
\end{table}

From Table~\ref{table:normal_quant}, it is evident that \texttt{MXINT8} and \texttt{INT4-VSQ}, which employ fine-grained per-block scale factors, result in significantly higher image quality compared to their INT8 and INT4 counterparts, which employ coarse-grained per-channel scale factors. 
Notably, the \texttt{MXINT8} configuration exhibits negligible degradation in image quality across all datasets. 
However, even with fine-grained scaling, the \texttt{INT4-VSQ} setting suffers from a substantial loss of image quality. 
While our exploration motivates the necessity of fine-grained scaling, it also suggests the need for new optimizations of the model and the data format to achieve baseline-level quality in low-precision.
In the following sections, we assume fine-grained scaling exclusively for data formats.

\subsection{Mixed-precision Quantization}

\begin{figure}[t]
    \centering
    \includegraphics[width=0.9\linewidth]{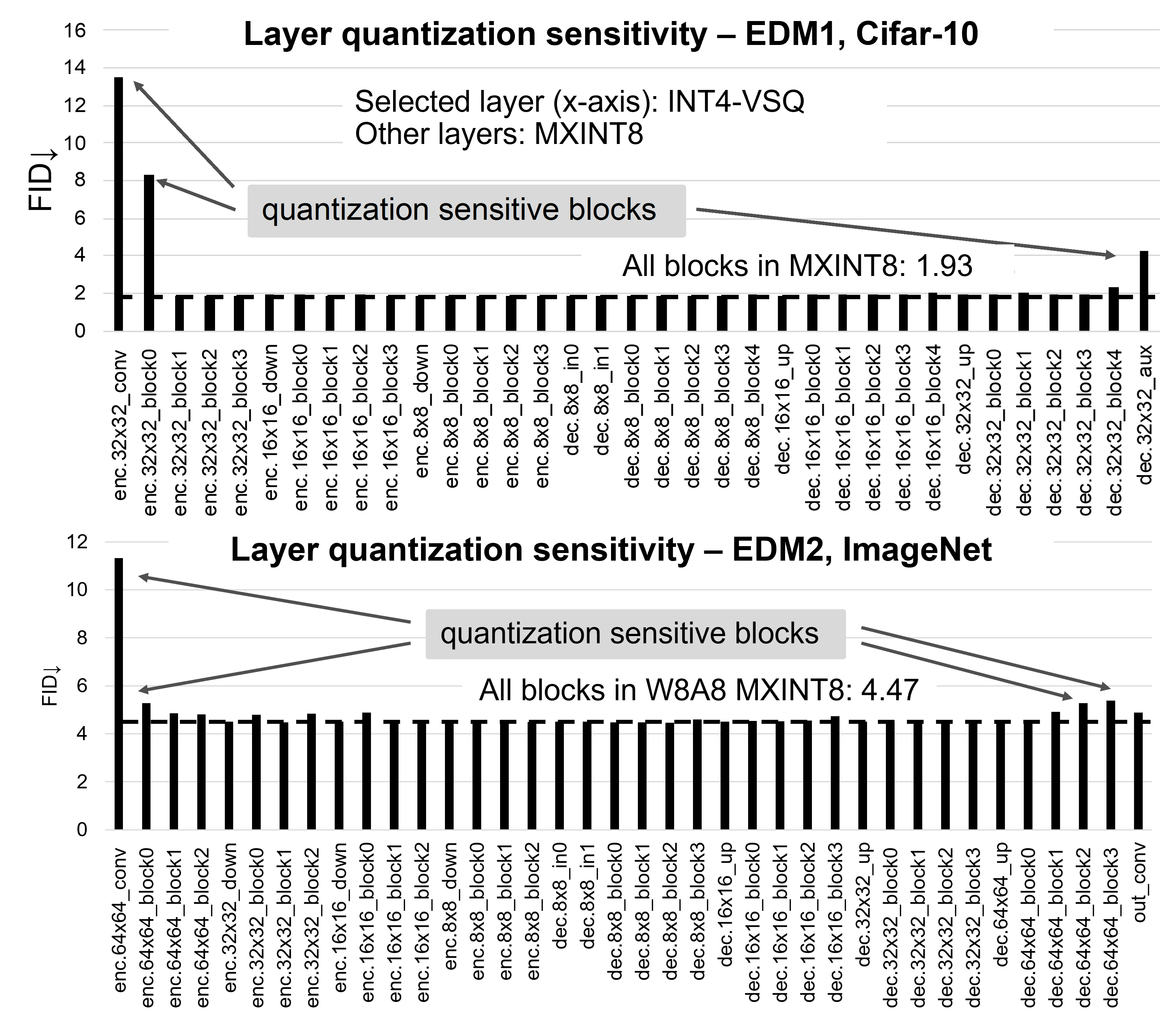}
    \caption{Block-wise quantization sensitivity for EDM model. }
    \label{fig:layer_sensitivity}
\end{figure}

The sensitivity to quantization across different layers or computation types is especially notable in these diffusion models, based on our block-wise quantization sensitivity experiment in Figure~\ref{fig:layer_sensitivity}.
In this experiment, we keep one block in 4-bit precision while all other blocks are set to 8-bit.
While we simply use \texttt{MXINT8} for the 8-bit blocks, we specifically propose our own \texttt{INT4} format with \texttt{FP8} scale factors for the 4-bit blocks to improve dynamic range of the representation.
The results in Figure~\ref{fig:layer_sensitivity} indicate that only the first and last few blocks are generally more sensitive to quantization. 
Consequently, it suffices to maintain these quantization-sensitive blocks at higher precision (\texttt{MXINT8}) while using 4-bit for the remaining blocks. 
The computation and memory cost of the high-precision blocks account for only about 5\% of the total cost, allowing the system to retain significant benefits from the 4-bit precision.

\begin{figure}[b]
    \centering
    \includegraphics[width=0.9\linewidth]{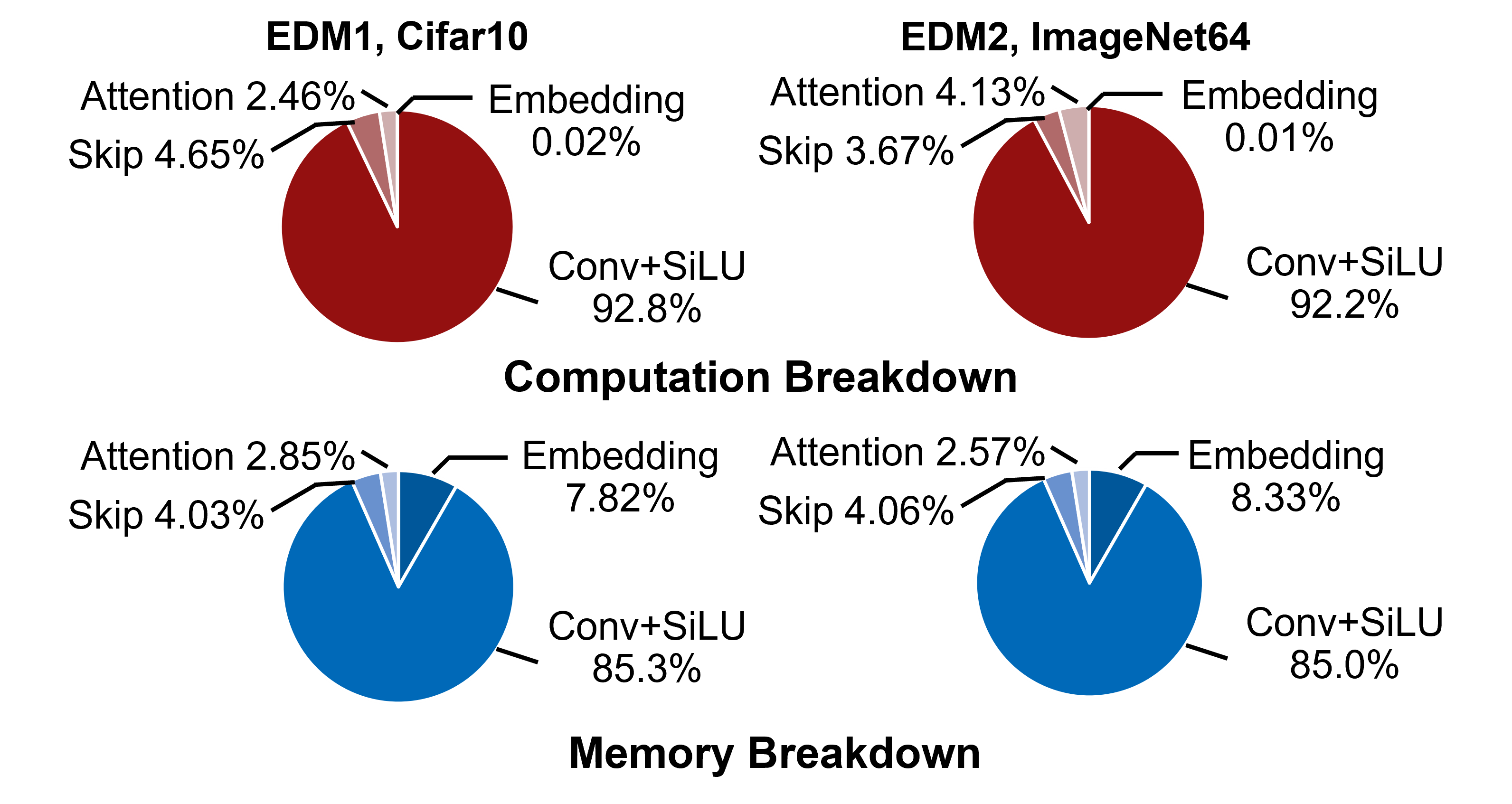}
    \caption{EDM model computation and memory breakdown. }
    \label{fig:computation_memory_breakdown}
\end{figure}

Additionally, Figure~\ref{fig:computation_memory_breakdown} presents a breakdown of the computation and memory costs for the four types of blocks in the model.
Notably, more than 90\% of the total computation cost and 85\% of the total memory cost are attributed to the \textit{Conv+SiLU} block, highlighting its significance in the model's overall performance. 
Therefore, we emphasize quantizing the \textit{Conv+SiLU} computation blocks to 4-bit, with other less important blocks in 8-bit.
The result of these optimizations are presented in the \texttt{Ours(MP-only)} row of Table~\ref{table:quant_exp}, demonstrating appreciable improvement from those of the baseline data format while attaining 73\% and 72\% average reduction in both computation and memory cost, respectively.
Here we assume a computational equivalence of 1 \texttt{FP16} multiplication to 2 \texttt{INT8} multiplications to 4 \texttt{INT4} multiplications based on computation resources and memory bandwidth~\cite{tirumala2024nvidia}.

\subsection{Hardware-efficient Activation Function}

\begin{figure}[t]
    \centering
    \includegraphics[trim={20pt 0pt 0pt 0pt}, clip, width=0.99\linewidth]{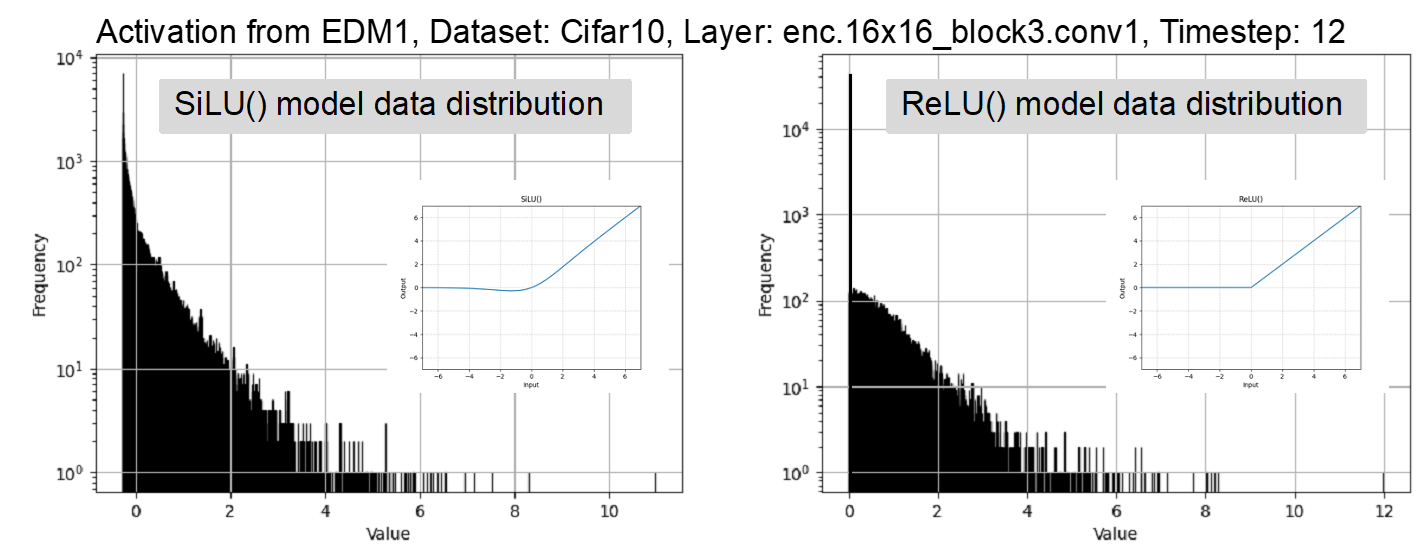}
    \caption{Comparison of activation data distributions at the output of Conv+SiLU versus Conv+ReLU. }
    \label{fig:silu_relu_distribution}
\end{figure}

\begin{figure}[t]
    \centering
    \includegraphics[width=0.9\linewidth]{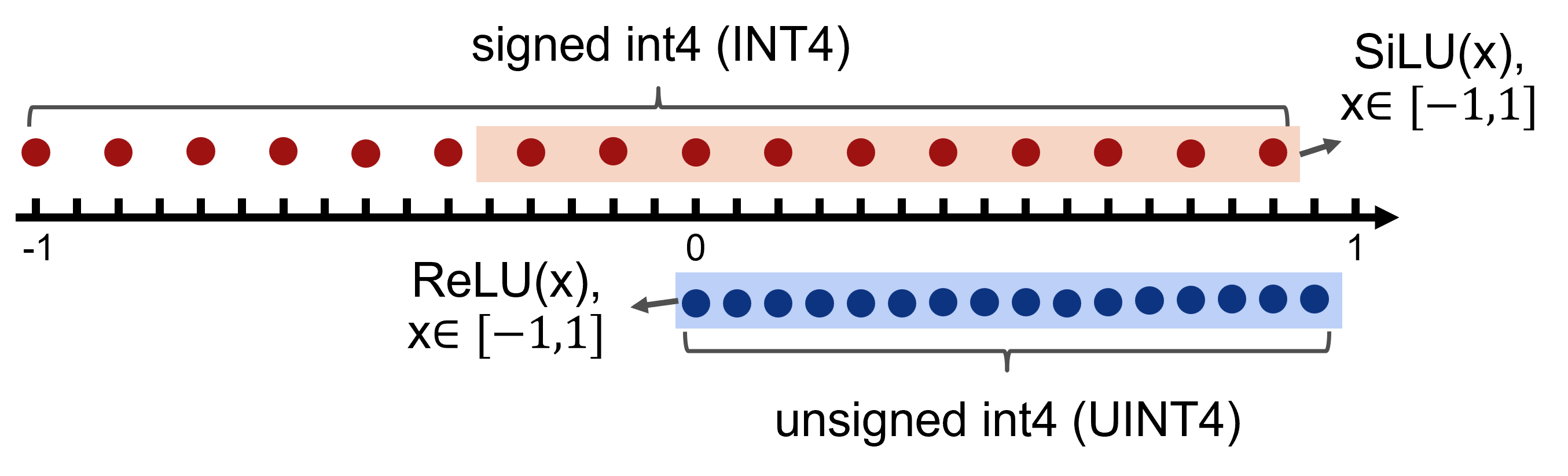}
    \caption{SiLU(x) after \texttt{INT4} quantization versus ReLU(x) after \texttt{UINT4} quantization.}
    \label{fig:int4_uint4}
\end{figure}

Although our mixed-precision quantization technique significantly improves the FID score, we identify additional optimization opportunities with the non-linear activation functions (SiLU) prevalent in these models.
Figure~\ref{fig:silu_relu_distribution} (left) shows the activation data distribution of one \textit{Conv + SiLU} layer where $\text{SiLU}(x) = x / {(1 + e^{-x})}$ \cite{elfwing2018sigmoid}. 
At the output of the SiLU non-linear function, the data distribution spans from \([-0.278, \infty)\). 
The existence of a small negative range necessitates the use of signed data formats (e.g., signed \texttt{INT4}) for activations. 
Figure~\ref{fig:int4_uint4} (top) illustrates the signed \texttt{INT4} quantization levels needed when SiLU is used. 
When the input \(x\) is in the range \([-1, 1]\), the output of \(\text{SiLU}(x)\) lies within \([-0.269, 0.731]\). 
Based on the quantization formula in Section~\ref{sec:quant}, only 10 of the 16 levels in signed \texttt{INT4} can be used, resulting in severe under-utilization of the available bitwidth.

In contrast, ReLU is a widely adopted and hardware-efficient non-linear function where the output data distribution ranges from \([0, \infty)\)~\cite{krizhevsky2012imagenet}. 
With ReLU in place of SiLU, we can focus our representation on only the positive range and quantize the activations to unsigned \texttt{INT4} (\texttt{UINT4}) data format to achieve more precise representations. 
Figure~\ref{fig:int4_uint4} (bottom) depicts the quantization of ReLU activations when \(x\) is within \([-1, 1]\). 
All the available quantization levels of the \texttt{UINT4} format can be used. 
As a result, leveraging \textit{Conv+ReLU} makes the model more quantization-friendly compared to using \textit{Conv+SiLU}.

To adapt the SiLU-based model to use ReLU, we replace the non-linear activation function and then finetune the pre-trained SiLU-based model. 
The finetuning process takes less than 10\% of the total pre-training time \cite{karras2022elucidating}. 
The resulting ReLU-based model achieves similar image quality to the original SiLU-based model. 
Figure~\ref{fig:image_example} presents example images (targeting AFHQv2 and FFHQ datasets) generated by the SiLU-based models with existing quantization techniques (left three images) and ReLU-based models (the fourth image) with our techniques, as well as their respective performance.
We can see that the 4-bit quantized ReLU-based model is clearly superior in quality to other 4-bit quantized models.

To derive the ReLU-based model, we finetune the full-precision version of the model followed by post-training quantization (PTQ) rather than directly performing quantization-aware training (QAT). 
This methodology avoids the overhead of QAT and produces a single ReLU-trained model that can be adapted to different quantization settings and hardware targets.
However, we do expect QAT on the ReLU-based model to attain incremental quality gain.
The FID score for the ReLU-based model with our proposed 4-bit quantization scheme is reported in the \texttt{Ours(MP+ReLU)} row in Table~\ref{table:quant_exp}. 
Indeed, by replacing the \textit{Conv+SiLU} block with the \textit{Conv+ReLU} block, our 4-bit diffusion model achieves the best FID scores.

\begin{table}[t]
    \centering
    \caption{FID comparison of different quantized models.}
    \label{table:quant_exp}
    \begin{tabularx}{0.5\textwidth}{>{\centering\arraybackslash}m{1.4cm}|>{\centering\arraybackslash}X|>{\centering\arraybackslash}X|>{\centering\arraybackslash}X|>{\centering\arraybackslash}X|>{\centering\arraybackslash}X|>{\centering\arraybackslash}X}
        \toprule
        \textbf{Quant Method} & \textbf{Avg. Comp. Saving} & \textbf{Avg. Mem. Saving} & \textbf{EDM1, CIFAR-10} & \textbf{EDM1, AFHQ v2} & \textbf{EDM1, FFHQ} & \textbf{EDM2, ImageNet} \\
        \midrule
        INT4-VSQ & 75\% & 75\% & 15.60 & 18.35 & 30.52 & 20.27 \\
        \hline
        Ours (MP-only) & 73\% & 72\% & 2.87 & 2.39 & 5.41 & 8.75 \\
        \hline
        \textbf{Ours (MP+ReLU)  } & \textbf{73\%} & \textbf{72\%} & \textbf{2.12} & \textbf{2.35} & \textbf{3.10} & \textbf{6.93} \\
        \bottomrule
    \end{tabularx}
\end{table}

\subsection{Temporal Per-channel Sparsity}
In addition to enabling aggressive 4-bit quantization, the use of ReLU simultaneously promotes significant activation sparsity in the model, as ReLU clamps all negative values strictly to zero. 
The average sparsity of the SiLU-based model is around 10\%, whereas the ReLU-based model achieves a significantly higher average sparsity of 65\% and up to 85\% sparse for some layers. 
However, mid-level random sparsity does not translate well to hardware acceleration; as shown in \cite{shinn2023sparsity}, unstructured sparsity requires at least 87.5\% sparsity to yield notable speed-ups on GPUs.
\begin{figure}[b]
    \centering
    \includegraphics[width=1.0\linewidth]{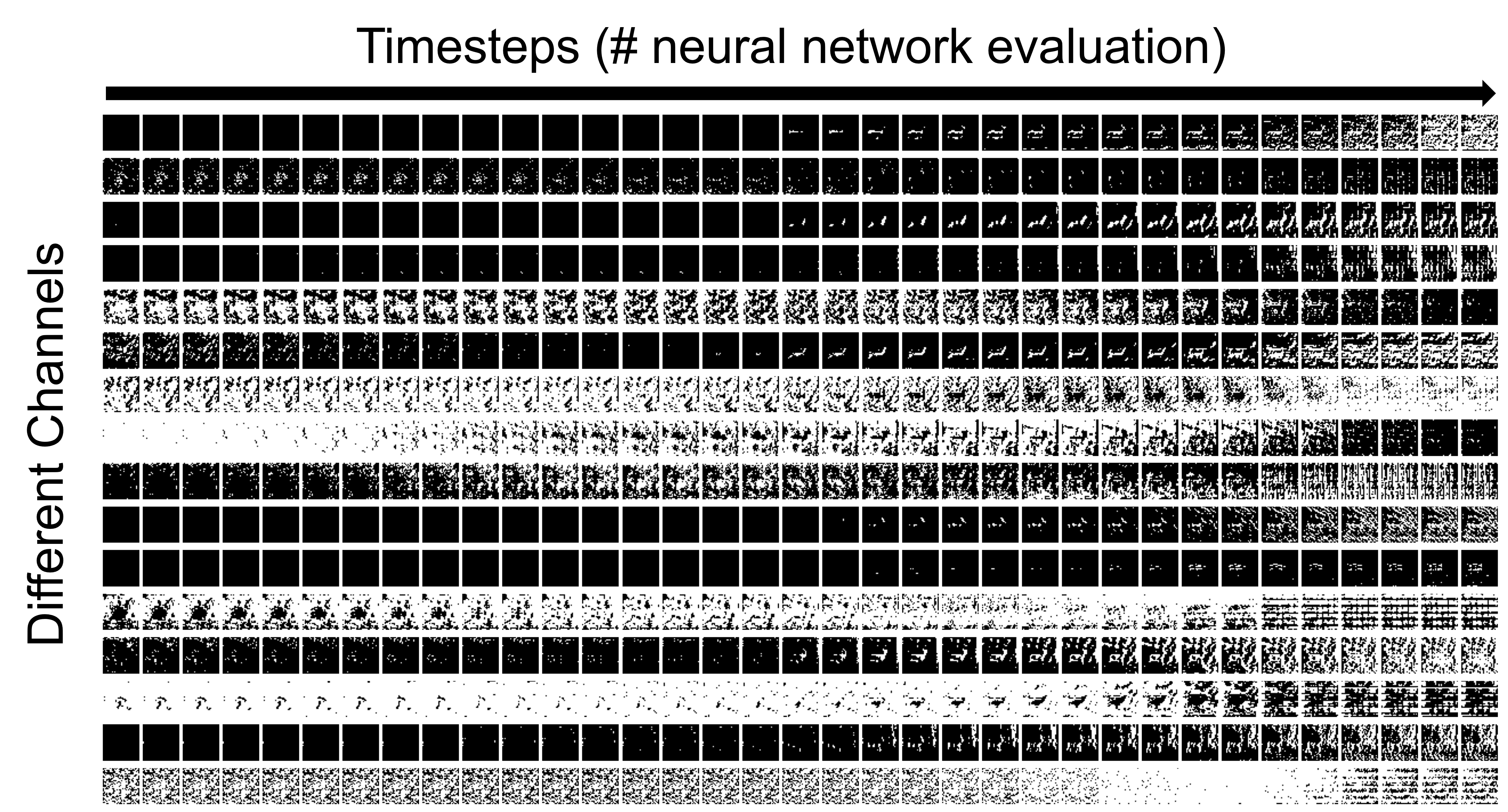}
    \caption{Temporal per-channel sparsity pattern.}
    \label{fig:sparsity}
\end{figure}

Interestingly, instead of random sparsity, we observe a temporal per-channel sparsity pattern in ReLU-based diffusion models that can be effectively exploited for further acceleration. 
Figure~\ref{fig:sparsity} depicts the sparsity pattern of the activations of a single layer in a ReLU-based EDM (for CIFAR-10 dataset). 
The values are binarized: zero values are represented in black, and non-zero values in white. 
Each row corresponds to a channel (32x32 block) of the activation tensor, while each column represents a time step in the diffusion process.

\begin{figure}[t]
    \centering
    \includegraphics[width=0.7\linewidth]{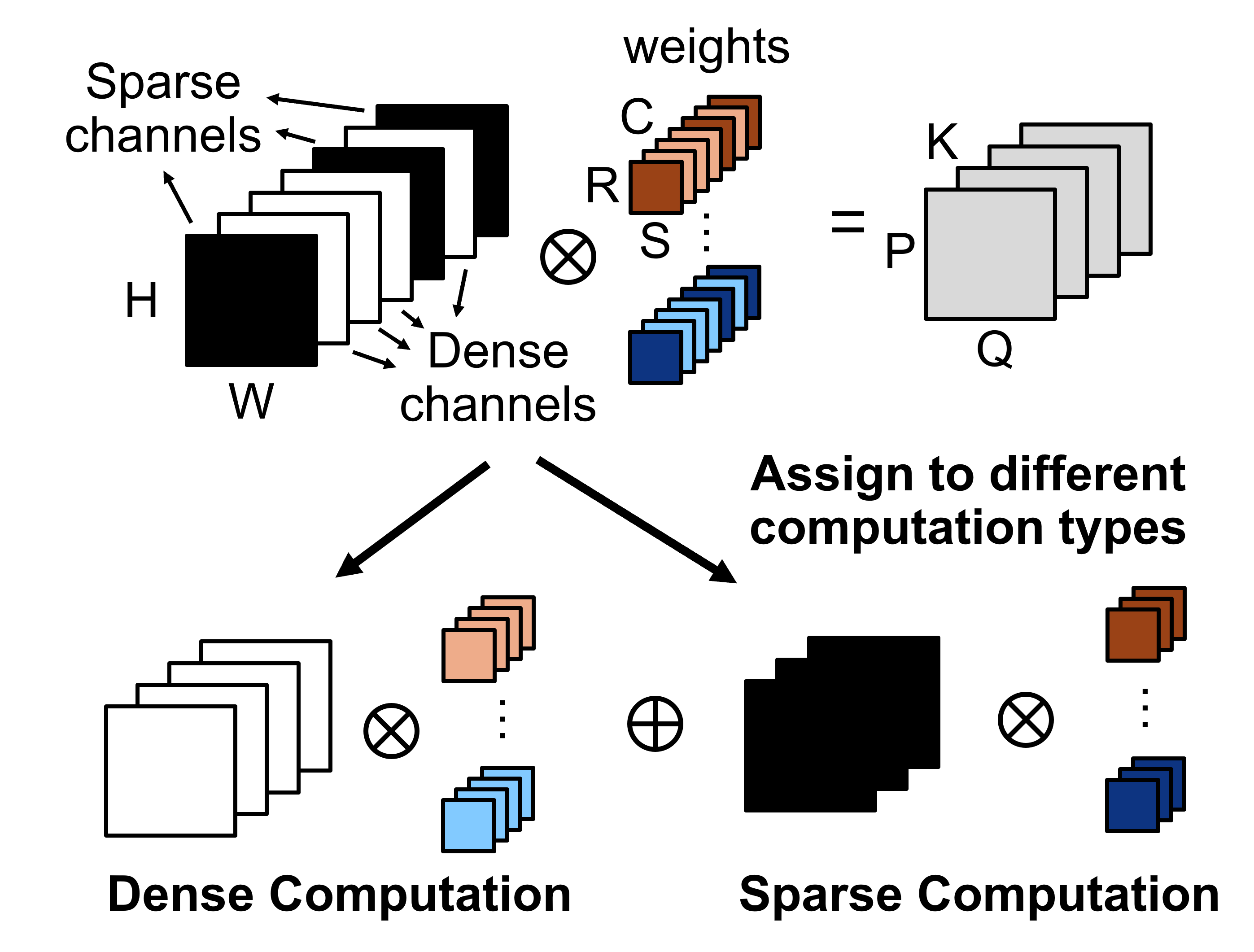}
    \caption{New computation scheme for temporal per-channel sparsity pattern.}
    \label{fig:computation}
\end{figure}

\begin{figure*}[bth]
    \centering
    \includegraphics[width=0.9\linewidth]{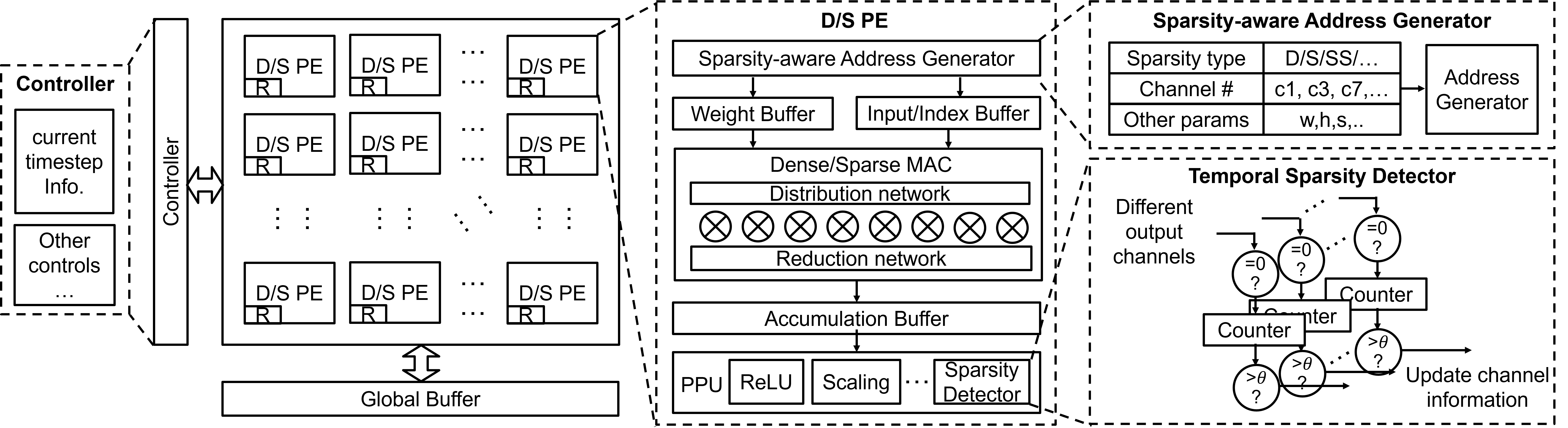}
    \caption{Overall architecture of our diffusion model accelerator.}
    \label{fig:overall_arch}
\end{figure*}

From Figure~\ref{fig:sparsity}, it is evident that different channels exhibit distinct sparsity characteristics; some channels are highly sparse (with predominantly black pixels) while others are dense (with mostly white pixels). 
Furthermore, the sparsity of each individual channel varies across time steps; sparse channels can become dense and vice versa. 
This temporal per-channel sparsity presents a perfect opportunity for acceleration: sparse channels can be grouped and processed using a sparse engine, while dense channels can be handled by a dense engine. 
In the following section, we introduce a new hardware architecture to support our aggressively quantized and temporal sparse diffusion models.

\section{Diffusion Model Accelerator}
\label{sec:hardware}
In this section, we propose a novel computation scheme and hardware architecture to accelerate the temporal per-channel sparsity pattern observed in ReLU-based model activations. Figure~\ref{fig:computation} illustrates the core concept: the activation tensor is categorized into sparse and dense channels. 
By grouping channels (including both weights and activations) based on the activation sparsity type, the computation can be optimized. 
Please note that the weights are always dense in this work. 
Under the proposed scheme, dense channel groups are processed using the dense processing unit, while sparse channel groups are computed using the sparse processing unit. 
After that, the partial sums are added together to get the final result. 
In following sections, we detail the architecture that leverages temporal sparsity to accelerate diffusion model while significantly reducing total system energy consumption.

\subsection{Overall Architecture}
Figure~\ref{fig:overall_arch} illustrates the overall architecture of our diffusion model accelerator design, adapted from the MAGNet modular accelerator generator~\cite{venkatesan2019magnet}. 
The accelerator comprises of three main components: a controller, a dense/sparse processing element (D/S PE) array, and an interconnection network between the global buffer and the PEs. 
The controller manages time step information and orchestrates various PEs through control logic.
The architecture features two types of PEs: Dense Processing Elements (DPEs) for dense channel computation and Sparse Processing Elements (SPEs) for sparse channel computation. 
These PEs are interconnected via configurable routers (R), enabling efficient processing of both dense and sparse data. 

The D/S PE consists of several key components: a sparsity-aware address generator, weight/input/accumulation buffers, dense/sparse datapaths, and a post-processing unit (PPU) with a sparsity detector. 
Each PE can be configured to either the dense or sparse datapath, depending on the computation type. 
Based on the sparsity pattern, either the dense or sparse vector MAC datapath is employed to compute partial sums. 
Recent works on dense accelerators \cite{jouppi2017datacenter, kwon2018maeri} and sparse accelerators \cite{han2016eie, zhang2016cambricon, gondimalla2019sparten, hegde2019extensor, qin2020sigma, srivastava2020tensaurus, shin2022griffin, qin2022enabling} have shown the effectiveness of tailored architectures for specific data types. 
In our design, we leverage a MAERI-like architecture~\cite{kwon2018maeri} for the dense MAC datapath and a SIGMA-like architecture~\cite{qin2020sigma} for the sparse MAC datapath, as both architectures are well-suited for handling irregular matrix sizes. 
SIGMA’s flexible distribution and reduction networks also enable efficient processing of irregular matrix sparsity. 

After processing all input channels, the data from the accumulation buffer is passed through the PPU. 
The sparsity-aware address generator maintains channel information, including sparsity type (dense, sparse, etc.) and the corresponding channel index. 
Utilizing this information, the address generator produces the necessary weight and activation addresses to fetch data from the global buffer. 
A temporal sparsity detector in the PPU identifies per-channel sparsity in the output. 
The number of zeros is compared against a predefined threshold to update the sparse channel index information for the next layer in the sparsity-aware address generator.

\subsection{Channel-last Memory Mapping}
\begin{figure}[t]
    \centering
    \includegraphics[width=1.0\linewidth]{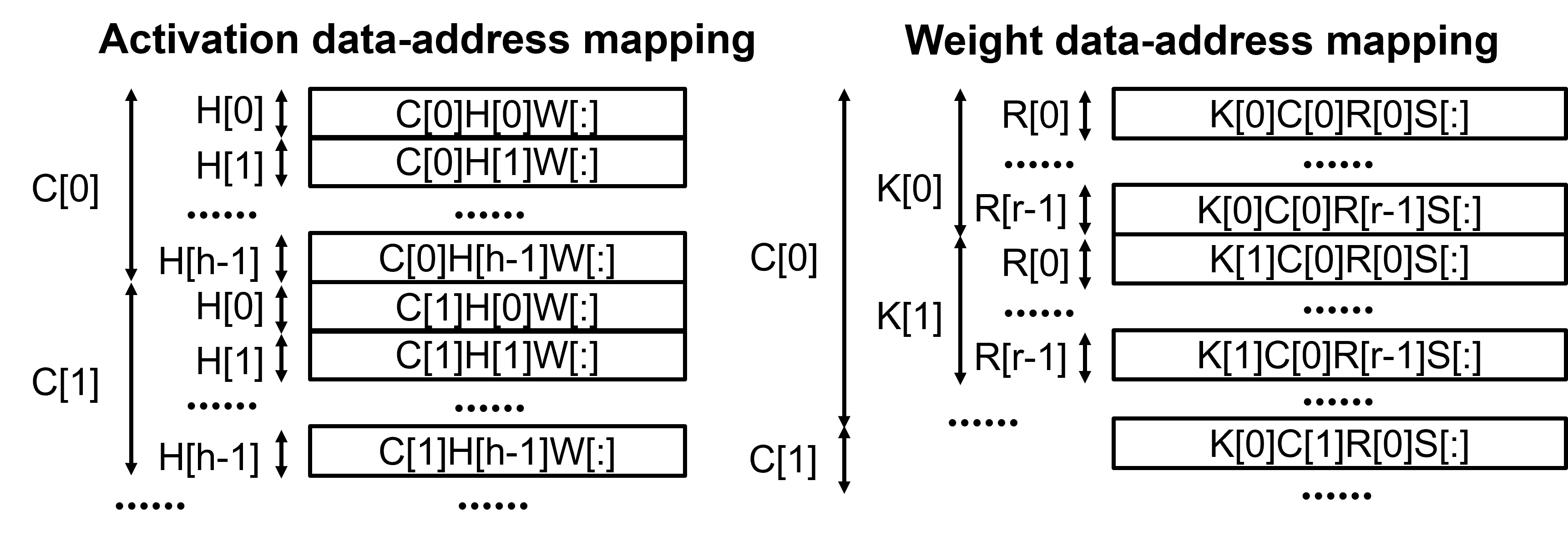}
    \caption{Channel-last data-address mapping.}
    \label{fig:address_mapping}
\end{figure}

To accommodate the non-continuous input channel order required by the sparsity-aware address generator when fetching data from the global buffer, we design a channel-last data mapping strategy, illustrated in Figure~\ref{fig:address_mapping}. 
For activations, the address mapping follows the sequence of width (W), height (H), and input/output channel (C) being the last, enabling the PE to fetch the entire dense/sparse input/output channel. 
For sparse channels, only nonzero values and its binary indicator (0 for zero and 1 for non-zero) are stored in the memory, which aligns with the SIGMA sparse accelerator architecture. 
For weights, the mapping sequence is kernel width (S), kernel height (R), output channel (K), and then input channel (C). 
This channel-last ordering ensures that the corresponding weights are fetched to align with the activations, optimizing the computation process.

\subsection{Temporal Sparsity Detection}
\begin{figure}[t]
    \centering
    \includegraphics[width=1.0\linewidth]{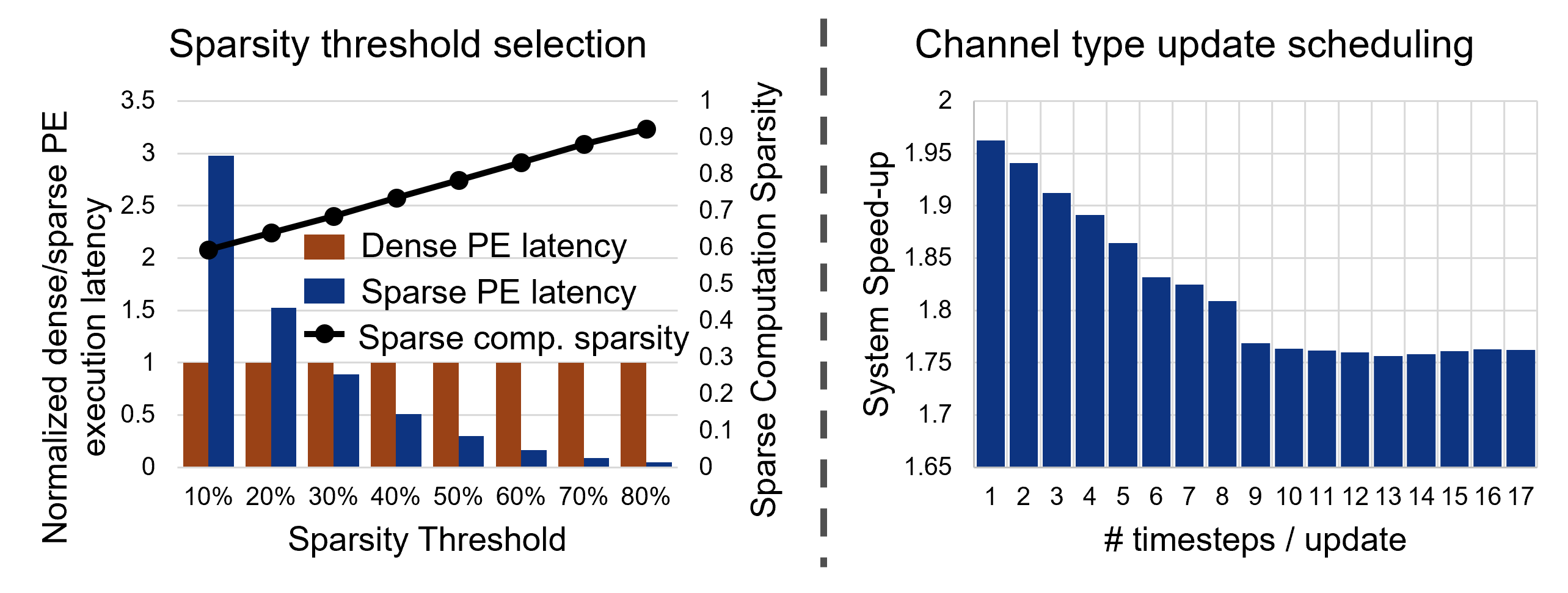}
    \caption{Analysis for temporal sparsity detection.}
    \label{fig:sparsity_update}
\end{figure}

The temporal sparsity detector calculates the output per-channel sparsity and assigns each channel as either dense or sparse. 
The sparsity threshold distinguishing dense from sparse channels is determined to balance the execution time between the dense PE and sparse PE. Figure~\ref{fig:sparsity_update} (left) analyzes the sparsity threshold. 
Based on this analysis, we select 30\% as the sparsity threshold. 
This threshold achieves an average sparsity of 70\% for the sparse tensor portion while maintaining a balance between the workloads of the dense and sparse PEs.

Finally, we propose a temporal sparsity update scheduling mechanism to address the dynamic per-channel sparsity observed across time steps. 
Figure~\ref{fig:sparsity_update} (right) illustrates the relationship between sparsity update frequency and system speed-up. 
More frequent sparsity updates (i.e., fewer time steps in between updates) improves the accuracy of sparse/dense tensor categorization, resulting in higher system speed-up.
Since the overhead of sparsity updates is negligible compared to the overall computation cost and can be hidden behind computation latency, we choose to update the per-channel sparsity at every time step during diffusion model execution to maximize categorization accuracy and system speed-up.


\subsection{Hardware Evaluation}
\begin{figure}[t]
    \centering
    \includegraphics[width=1.0\linewidth]{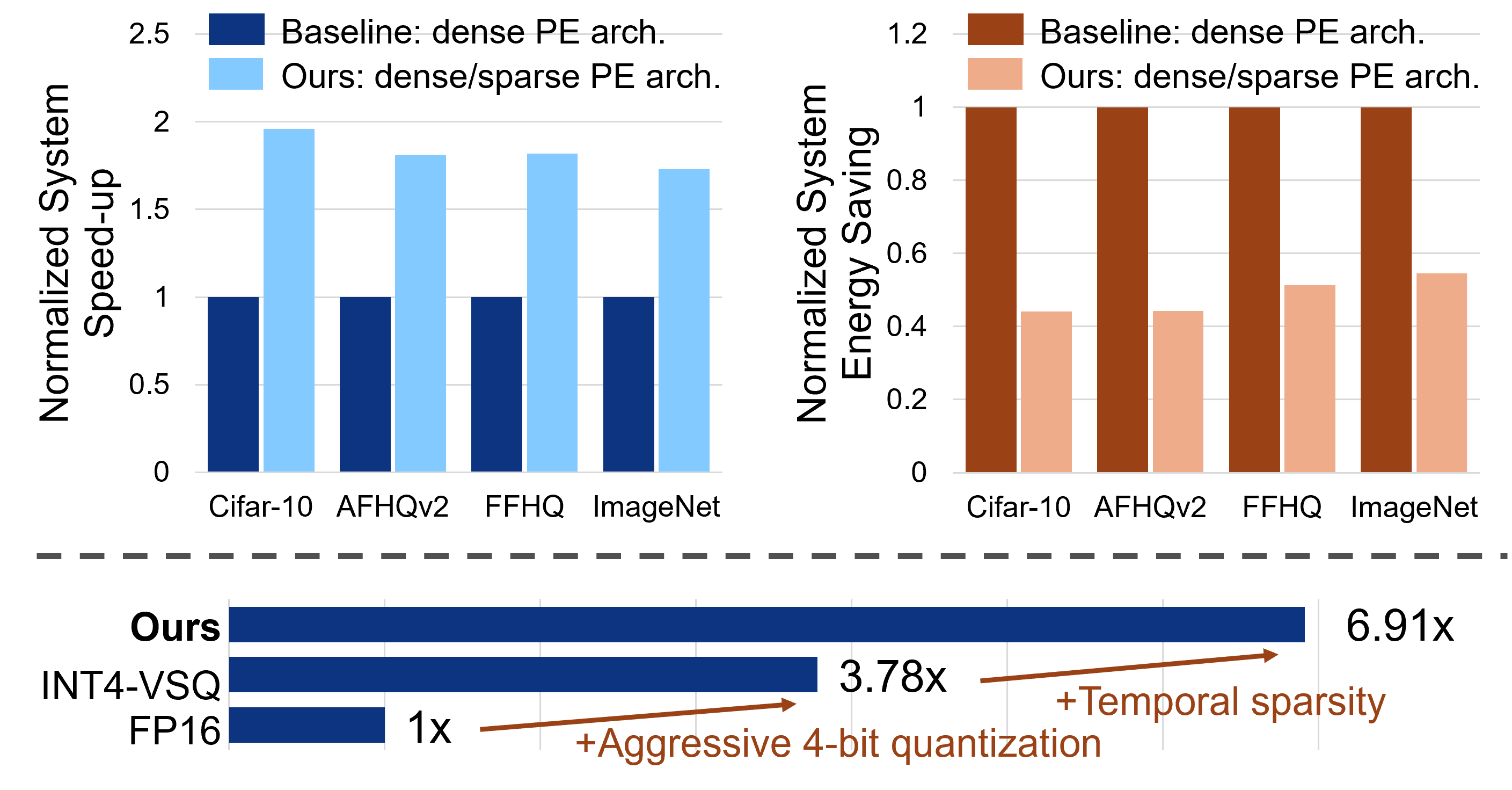}
    \caption{System evaluation.}
    \label{fig:system_evaluation}
\end{figure}

We utilize Stonne~\cite{munoz2021stonne}, an open-source simulation framework, to evaluate the latency and energy consumption of the DPE and SPE. 
Stonne provides end-to-end evaluation of flexible accelerator microarchitectures with sparsity support. 
The design is simulated under 28nm technology. 
In these experiments, we assume the architecture includes one DPE and one SPE, each containing 128 multipliers. 
This architecture is scalable to meet specific latency and power requirements. 
The baseline for comparison is a purely dense architecture with two DPEs. 
Figure~\ref{fig:system_evaluation} (top) presents the average speed-up and energy saving across different datasets relative to the baseline. 
Focusing solely on temporal sparsity in this figure, we achieve an average speed-up of \(1.83\times\) along with a system energy saving of 51.5\%.

Figure~\ref{fig:system_evaluation} (bottom) illustrates the total speed-up compared to an FP16 SiLU-based diffusion model. 
Our 4-bit quantization contributes to \(3.78\times\) speed-up, while our temporal per-channel sparsity adds \(1.83\times\) speed-up on top of quantization. 
In combination, our proposed model optimizations and hardware architecture described in this paper achieve a total speed-up of \(6.91\times\).

\section{Conclusions}

\label{sec:conclusions}

In this work, we propose a set of co-designed optimization techniques to aggressively quantize diffusion models to 4-bit while simultaneously promoting significant activation sparsity. 
By designing a novel heterogeneous dense/sparse accelerator architecture, we achieve a \(6.91\times\) speed-up compared to an FP16 baseline while demonstrating state-of-the-art image generation quality. 
By leveraging temporally sparse computations, we save 51.5\% in energy consumption compared to traditional dense accelerators. 
In the future, we plan to extend our techniques to diffusion models targeting video generation~\cite{ho2022video} and apply our methodology to other generative models. 
Exploring new DNN models will allow us to further enhance our optimization techniques and expand the applicability of our proposed accelerator design.


\newpage
\bibliographystyle{ieeetr}
\bibliography{ref.bib}


\end{document}